\documentclass{article}
\usepackage[preprint]{neurips_2026}
\usepackage[utf8]{inputenc}
\usepackage[T1]{fontenc}
\usepackage{hyperref}
\usepackage{url}
\usepackage{booktabs}
\usepackage{amsfonts}
\usepackage{amsmath,amssymb}
\usepackage{nicefrac}
\usepackage{microtype}
\usepackage{graphicx}
\newcommand{\W}{W_2}
\title{Measuring Optimal Transport in Transformer Depth}
\author{%
  Alexandre Quemy\\
  Hother Labs\\
  \texttt{alexandre@hother.io}
}
\begin{document}
\maketitle

\begin{abstract}
A transformer carries each token's state from layer to layer, and the whole vocabulary carried together forms a cloud that moves with depth. We ask whether a trained network moves this cloud the way optimal transport would: at the cheapest cost, and along the map that pairs each token with its optimal destination. We measure both on Pythia-160m and Pythia-410m, with an exact assignment between consecutive layer clouds, a measured sampling floor, calibration on couplings known to be optimal, and a split of the cost into the common shift of the cloud and the token-specific moves. At the last layer, both models move their tokens where the optimal-transport map sends them, at the optimal cost for Pythia-410m and slightly above it for Pythia-160m. At the first layer they do not. In between, single layers can be judged on cost at only two of ten transitions, and blocks of several layers move the cloud at close to the optimal cost. The agreement at the last layer is much weaker at initialisation (0.64 against 0.86) and grows with training. 
\end{abstract}

\section{Introduction}
A transformer with $L$ layers turns each token vector $x^{(0)}$ into $x^{(1)},\dots x^{(L)}$. 
At layer $0$ the state is the embedding of the current token, so there are as many distinct points as the vocabulary cardinality.
From layer $1$ on, states depend on context, so there are as many points as token occurrences. 
Taken together, the states of a corpus form a cloud $\mu_\ell$ at layer $\ell$. The network moves the first cloud onto the second with a specific pairing, token $i$ goes from $x_i^{(\ell)}$ to $x_i^{(\ell+1)}$.

Optimal transport asks what the cheapest way to move $\mu_\ell$ onto $\mu_{\ell+1}$ would be if the pairing were free, paying squared distance per point:
\begin{equation}
\W^2(\mu_\ell,\mu_{\ell+1}) = \min_{\pi} \ \mathbb{E}_{(x,y)\sim\pi}\,\|x-y\|^2 .
\end{equation}
For this cost, the optimal pairing is unique and is the gradient of a convex function, $y=\nabla\varphi(x)$ \citep{brenier1991polar}.

In this paper, we ask whether the network's own pairing is that particular one. To do so, we study the transport efficiency of Pythia-160m and Pythin 410M, layer per layer, over a sample of token positions obtained on a subset of WikiText-103 and Piles datasets. 
A following question we study is whether the move of token $i$ from layer $\ell$ to $\ell{+}1$ matches the optimal-transport displacement~$\nabla\varphi_\ell(x_i)-x_i$.

There are several practical interests in answering these questions. If a layer follows that map, its action on all tokens is summarised by a single convex potential, so the layer can be replaced by a (possibly) cheap optimal-transport surrogate for analysis \citep{makkuva2020optimal,korotin2023neural}, and the methods that merge models \citep{singh2020fusion} or align their representations \citep{alvarezmelis2018gromov} with optimal transport can be applied with the guarantee that the network's own map is of the same kind. In addition, the distance to the cheapest move is a label-free measure of wasted motion \citep{finlay2020train,kan2025ottransformer}, usable to compare layers and models, and the agreement with the optimal map across training checkpoints \citep{biderman2023pythia} shows when and where a layer's function forms.


We make three contributions. The first is the question itself. To our knowledge, this is the first empirical test of whether the layer-to-layer map of a trained transformer acts as optimal transport between its own consecutive layer clouds, using the coupling the network defines, both in cost and in the pairing it induces.
 The second is a measurement protocol for optimal transport between layer transitions: the sampling error is measured and subtracted, the readings are calibrated on couplings known to be optimal, the common shift of the cloud is separated from the token-specific moves, and a lower bound that needs no calibration is provided. The third is the result: on Pythia-160m and Pythia-410m, the last layer moves its tokens along the optimal-transport map, this agreement appears with training, and the cost of the network's moves is optimal wherever it can be measured, with one exception at the exit of the smaller model.

The rest of the paper is organized as follows. Section~\ref{sec:related} discusses related work. Section~\ref{sec:question_1} studies the cost of the network's moves and Section~\ref{sec:question_2} their agreement with the optimal-transport map. We conclude in Section~\ref{sec:conclusion}.

\section{Related work}\label{sec:related}
\textbf{Transformers as maps between measures.} The literature models tokens as interacting particles \citep{vuckovic2020mathematical}, their clustering in the long-time limit \citep{geshkovski2025mathematical}, and transformers as measure-to-measure maps with a Vlasov limit \citep{furuya2025support}. We share the object, the cloud of token states carried from layer to layer, but these works derive properties of the map in theory, whereas we measure on trained networks whether the map is the cheapest one, in cost and in the pairing it induces.

\textbf{Attention and optimal transport.} Doubly stochastic attention is read as a Wasserstein gradient flow \citep{sander2022sinkformers}, and attention formulated as semi-relaxed entropic OT inside a layer, with GPT-2 drift profiles \citep{anon2025transformersot}. Both place optimal transport inside the attention operation, at a fixed regularisation. We place it between consecutive layer clouds, unregularised, and test the network's coupling against the exact optimal one; neither work contains an efficiency or map test.

\textbf{OT-regularised flows.} OT-Transformer \citep{kan2025ottransformer} and rectified flows \citep{liu2023rectified} add a kinetic-energy penalty so that the learned flow takes optimal paths. They impose the property we measure: our question is whether a transformer trained with no such penalty sits near the optimal coupling anyway.

 \textbf{Depth as dynamics.} Linear-drift SDEs fitted to layer trajectories \citep{sarfati2025lines} and linear cross-layer lenses \citep{belrose2023tuned} describe depth by a fitted linear map. Our linear baseline $T_G$ is the optimal-transport member of that family, and our control asks what the network does beyond it: at the 160m exit one third of the move is beyond any linear map, and 61\% beyond the optimal linear map $T_G$, and still follows optimal transport.






\section{Transport efficiency}\label{sec:question_1}


When a layer moves the cloud of token states from layer $l$ to $l+1$, does it pay the smallest possible price for that change of cloud? 
 The smallest possible price is the optimal-transport cost between the two clouds:

$$e_\ell=\frac{W_2^2(\mu_\ell,\mu_{\ell+1})}{\mathbb E_i|x_i^{(\ell+1)}-x_i^{(\ell)}|^2},\qquad W_2^2(\mu_\ell,\mu_{\ell+1})=\min_{\pi}\ \mathbb E_{(x,y)\sim\pi}|x-y|^2 ,$$

the minimum being over all pairings $\pi$ of the two clouds. Since the network's pairing is one candidate, $e_\ell\le1$, and $e_\ell=1$ means the network is a perfect mover. Because a common shift of the whole cloud is exactly optimal (and in fact dominates most transitions), the question is asked on the token-specific part:

Write $d_i = x_i^{(\ell+1)}-x_i^{(\ell)}$ for token $i$'s move and $\Delta m_\ell=\mathbb E_i\, d_i$ for the mean move. Then
$$\bar e_\ell=\frac{W_2^2(\bar\mu_\ell,\bar\mu_{\ell+1})}{P_\ell},\qquad P_\ell=\mathbb E_i\|d_i-\Delta m_\ell\|^2,\qquad e_\ell=\frac{A_\ell+\bar e_\ell P_\ell}{A_\ell+P_\ell},$$

with $A_\ell=\|\Delta m_\ell\|^2$ the cost of the common shift.

 We use the two following objects throughout this paper: 
 \begin{itemize}
 \item The \emph{mean flow}: one point per token type, its residual-stream state averaged over all occurrences in a corpus. 
 \item The \emph{raw states}: one point per occurrence, the state at an actual position in an actual document.
 \end{itemize}

We studied Pythia-160m (12 layers) and Pythia-410m (24 layers)~\citep{biderman2023pythia} and used WikiText-103 and Pile as corpora. On WikiText, we collected 242{,}015 positions in the full 768-d space, i.e. raw per-occurrence states, leading to  25{,}268 token types for the corpus-mean token trajectories. 
On Pile, we collected 46{,}542 corpus-mean token trajectories only.

We then split the study into two: one in reduced dimension, and one the full space.


 \subsection{Exact efficiency in a 16-d camera}

\paragraph{Protocol}

 We sample 4,000 tokens at layer $\ell$ and an independent 4,000 at $\ell+1$, form $C_{ij}=\|x_i-y_j\|^2$ and solve the assignment exactly using the network simplex of POT~\citep{flamary2021pot}, the implementation of \citet{bonneel2011displacement}. This is the $\W^2$ of the two samples with no regularisation. 

 
 It is computed in a balanced 16-d PCA camera because the error of an empirical Wasserstein distance decays like $n^{-1/d}$ \citep{fournier2015rate,weed2019sharp}. 
 Note that because the two samples are independent, the numerator carries a finite-sample floor that the paired denominator does not. As a result, the raw ratio $e_\ell$ can exceed $1$. 

 To make the number honest, we subtract the assignment cost between two independent samples of the \emph{same} layer. This is the self-term correction that defines the Sinkhorn divergence \citep{feydy2019interpolating,chizat2020faster}, applied here to the exact assignment and to the target layer only. Where the floor exceeds $30\%$ of the cost in question, the transition is reported as unresolvable. It essentially means that the token-specific moves at that transition are smaller than the typical distance between neighbouring sampled tokens, so we cannot separate them from the sampling gap.

 In addition, to calibrate the reading, we feed the same pipeline two couplings that are optimal by construction: 1) a shift $y=x+t$ with $t=\Delta m_\ell$ and 2) a radius-matched expansion $y=s(x-m_\ell)+m_{\ell+1}$, built from independent samples of layer $\ell$. Their reading, $0.85$--$1.00$ across the resolvable transitions, is what $e_\ell=1$  looks like through our instrument.
 Efficiency is reported raw ($e_\ell$) and on the centered clouds ($\bar e_\ell$), with $A/P$. The whole protocol is repeated over 3 seeds.


\begin{table}[!t]\centering\scriptsize
\caption{Transport efficiency of Pythia-160m (16-d camera, $n=4{,}000$, three seeds). Each efficiency is shown next to what an optimal coupling reads through the same instrument; $\bar e^{\mathrm{sliced}}_\ell$ is the lower bound from all types; block rows pair layer $\ell$ with $\ell{+}k$ directly. n.r.: not resolvable.}
\label{tab:cost160}
\begin{tabular}{l c c cc cc c}
\toprule
 & & & \multicolumn{2}{c}{raw} & \multicolumn{2}{c}{token-specific} & \\
transition & floor / $P_\ell$ & $A/P$ & $e_\ell$ & optimal reads & $\bar e_\ell$ [seed range] & optimal reads & $\bar e^{\mathrm{sliced}}_\ell$ \\
\midrule
0$\to$1 & 0.08 & 3.8 & 0.98 & 1.00 & 1.02 [1.00--1.04] & 1.01 & 0.94 \\
1$\to$2 & 0.51 & 3.4 & 0.90 & 0.99 & n.r. & -- & 0.32 \\
2$\to$3 & 0.50 & 0.7 & 0.82 & 0.99 & n.r. & -- & 0.60 \\
3$\to$4 & 0.09 & 0.1 & 0.99 & 0.93 & 1.07 [1.03--1.15] & 1.13 & 0.78 \\
4$\to$5 & 2.73 & 1.2 & n.r. & -- & n.r. & -- & 0.36 \\
5$\to$6 & 2.92 & 5.3 & n.r. & -- & n.r. & -- & 0.22 \\
6$\to$7 & 2.47 & 1.4 & n.r. & -- & n.r. & -- & 0.18 \\
7$\to$8 & 0.73 & 3.7 & 0.90 & 0.94 & n.r. & -- & 0.21 \\
8$\to$9 & 0.59 & 6.3 & 0.92 & 0.96 & n.r. & -- & 0.24 \\
9$\to$10 & 0.58 & 3.1 & 0.86 & 0.94 & n.r. & -- & 0.29 \\
10$\to$11 & 0.13 & 1.5 & 0.94 & 0.99 & 0.99 [0.97--1.02] & 1.09 & 0.46 \\
11$\to$12 & 0.11 & 99.0 & 1.00 & 1.00 & 0.86 [0.85--0.87] & 1.00 & 0.73 \\
\midrule
3$\to$9, six layers as a block & 0.08 & 1.3 & 0.95 & 1.00 & 0.97 & 0.95 & 0.67 \\
8$\to$11, three layers as a block & 0.06 & 2.4 & 0.98 & 1.01 & 0.96 & -- & 0.48 \\
\bottomrule
\end{tabular}

\end{table}
\begin{table}[!t]\centering\scriptsize
\caption{ Transport efficiency of Pythia-410m (16-d camera, $n=4{,}000$, one seed), same columns as Table 1; $\bar e^{\mathrm{sliced}}_\ell$ from all 11,937 types. The block row pairs layer 4 with layer 20 directly, the only resolvable block of this model. Only the transitions resolvable on at least one cost are shown.}
\label{tab:cost410}
\begin{tabular}{l c c cc cc c}
\toprule
 & & & \multicolumn{2}{c}{raw} & \multicolumn{2}{c}{token-specific} & \\
transition & floor / $P_\ell$ & $A/P$ & $e_\ell$ & optimal reads & $\bar e_\ell$ & optimal reads & $\bar e^{\mathrm{sliced}}_\ell$ \\
\midrule
0$\to$1 & 0.06 & 4.3 & 0.98 & 0.98 & 1.01 & 1.01 & 0.99 \\
5$\to$6 & 0.02 & 0.2 & 1.11 & 0.99 & 0.97 & 1.09 & 0.83 \\
6$\to$7 & 0.09 & 0.2 & 1.24 & 1.17 & 0.93 & 0.92 & 0.99 \\
22$\to$23 & 0.08 & 0.5 & 0.87 & 1.23 & 1.02 & 0.93 & 0.85 \\
23$\to$24 (exit) & 0.05 & 1.6 & 0.98 & 0.98 & 0.99 & 1.09 & 0.49 \\
\midrule
4$\to$20, sixteen layers as a block & 0.02 & 1.0 & 0.97 & 1.00 & 0.96 & 1.03 & 0.86 \\
\bottomrule
\end{tabular}
\end{table}
\paragraph{Results}
We report the results in Tables~\ref{tab:cost160} and~\ref{tab:cost410}, respectively for Pythia-160m and Pythia-410m and in Table~\ref{tab:ap} the ratio $A/P$ in the full space next to its camera value, which shows how the camera changes that ratio.

A shift of the whole cloud is optimal for free, and it dominates the cost: $A/P$ is $1.5$--$2.5$ in the full space at most transitions and $84$ at the 160m exit. The camera changes $A/P$ by up to a factor of three either way, since it keeps different shares of the shift and of the token-specific movement at each transition (Table~\ref{tab:ap}), but the full-space values, computed without a camera, show the same behavior. Therefore, we are confidence that using the reduced space does not alter the general conclusions.

Once the common shift is removed, the network's move can be compared with the cheapest move properly. At three of the four measurable transitions in Pythia-160m (entry, 3→4, 10→11) and at every measurable transition in Pythia-410m, the network coupling is optimal, within the instrument's precision of 10–15\%. In particular, at the entry, the sliced bound says the true efficiency is at least 0.94, so the network is within 6\% of optimal there.


 One measured gap: the 160m exit. The token-specific efficiency reads 0.86 where the calibration reads 1.00. The network pays between 14\% and 27\% more than the cheapest move there. The 410m exit shows no gap. It remains an open question to know if this is a property of the model or the scale.

  Between layers 4 and 10 the token-specific moves are tiny compared with the distance between tokens. The token-specific moves there are too small for 4,000 samples to resolve, and the sliced bound only says the efficiency is above 0.2–0.4. No cost verdict either way at the single-layer scale. However, pairing layer $\ell$ directly with layer $\ell{+}k$ makes the accumulated step resolvable once it is large enough: six layers in the early middle (3→9 reads $\bar e = 0.97$ against a calibration of 0.95) and three in the late middle (8→11 reads 0.96). 

  Similarly, most transitions of Pythia-410m are not resolvable. With twice the depth, each layer moves its tokens less relative to their spacing (the token-specific movement is 3–6\% of the cost in the middle, against 16–32\% for Pythia-160m), and a larger sample would help only slowly. Our hypothesis is that those small moves are linear, as the 410m exit is to within resolution but testing it is left for future work.


\begin{table}[!t]\centering\scriptsize
\caption{$A/P$ in the full 768-d space and in the 16-d camera, Pythia-160m WikiText mean flow, with the share of the common-shift energy $A$ and of the token-specific energy $P$ that the camera keeps.}
\label{tab:ap}
\begin{tabular}{l cccccccccccc}
\toprule
transition & 0$\to$1 & 1$\to$2 & 2$\to$3 & 3$\to$4 & 4$\to$5 & 5$\to$6 & 6$\to$7 & 7$\to$8 & 8$\to$9 & 9$\to$10 & 10$\to$11 & 11$\to$12 \\
\midrule
$A/P$, full space & 1.5 & 1.5 & 0.7 & 0.3 & 1.0 & 1.8 & 1.0 & 1.8 & 2.5 & 2.1 & 1.6 & 84 \\
$A/P$, camera & 3.7 & 3.1 & 0.6 & 0.1 & 1.2 & 5.2 & 1.4 & 3.7 & 6.2 & 3.1 & 1.5 & 109 \\
$A$ kept by camera & 0.74 & 0.55 & 0.29 & 0.30 & 0.24 & 0.46 & 0.24 & 0.64 & 0.70 & 0.36 & 0.49 & 0.84 \\
$P$ kept by camera & 0.30 & 0.26 & 0.30 & 0.83 & 0.20 & 0.16 & 0.17 & 0.32 & 0.28 & 0.24 & 0.52 & 0.65 \\
\bottomrule
\end{tabular}
\end{table}

\subsection{A lower bound without calibration}

\paragraph{Protocol} We use the sliced Wasserstein construction~\citep{bonneel2015sliced}. For a random unit direction $u$, we project every point on $u$, sort, and compute the sorted cost $W_{2,u}^2$, which is the exact optimal transport in one dimension. Since $\mathbb{E}_u (u\cdot v)^2=\|v\|^2/D$ and sorting is the minimum over all pairings of the projected points, $D\,\mathbb{E}_u W_{2,u}^2 \leq \W^2(\mu_\ell,\mu_{\ell+1})$, with equality for a shift or a uniform scaling and a strict inequality in general. Averaging over $2{,}000$ directions on the centered clouds inside the camera, with all $25{,}268$ types on both sides and no sampling, gives
\begin{equation}
\bar e^{\mathrm{sliced}}_\ell \;=\; \frac{D\,\mathbb{E}_u W_{2,u}^2(\bar\mu_\ell,\bar\mu_{\ell+1})}{P_\ell} \;\leq\; \bar e_\ell ,
\end{equation}
a lower bound on the token-specific efficiency that needs no sampling, no floor and no calibration.

Because it is a lower bound, the true efficiency is at least $\bar e^{\mathrm{sliced}}_\ell$ (Tables~\ref{tab:cost160} and~\ref{tab:cost410}): $0.94$ at the entry and $0.73$ at the exit of Pythia-160m, and $0.86$ over the sixteen-layer block $4\to20$ of Pythia-410m, with no assumption.

In conclusion, wherever the cost can be measured, the network moves its tokens at the price of an optimal coupling, with one exception, the exit of Pythia-160m, which pays 14 to 27\% more.

\begin{figure}[t]\centering
\includegraphics[width=0.6\linewidth]{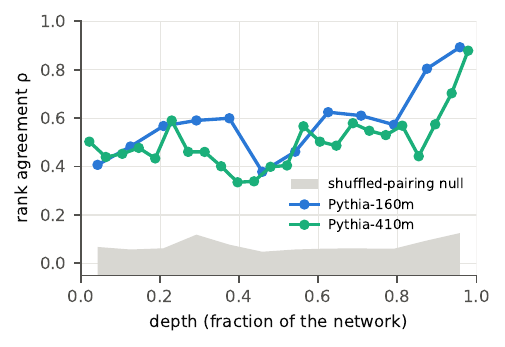}
\caption{Rank agreement $\rho$ between the network's move and the optimal-transport move, against depth, both models, 16-d camera, exact assignment on the centered clouds, with the shuffled-pairing null (grey). Agreement rises from the entry to the exit on both models.}
\label{fig:profile}\end{figure}


\section{Agreement with the Brenier map}\label{sec:question_2}

The second question is whether token $i$'s actual move matches the Brenier route $\nabla\varphi(x_i)-x_i$. 
On the centered clouds, the exact assignment sends each sampled token $x_i$ to the destination $y_{\sigma(i)}$ an optimal mover would choose. 
We compare the network's move $d_i$ with the optimal move $y_{\sigma(i)}-x_i$ by the Spearman rank agreement $\rho$ of the pooled components and by the median cosine between the two moves. 
Because both clouds are centered, a common shift cannot create agreement. 

\begin{figure}[t]\centering
\includegraphics[width=\linewidth]{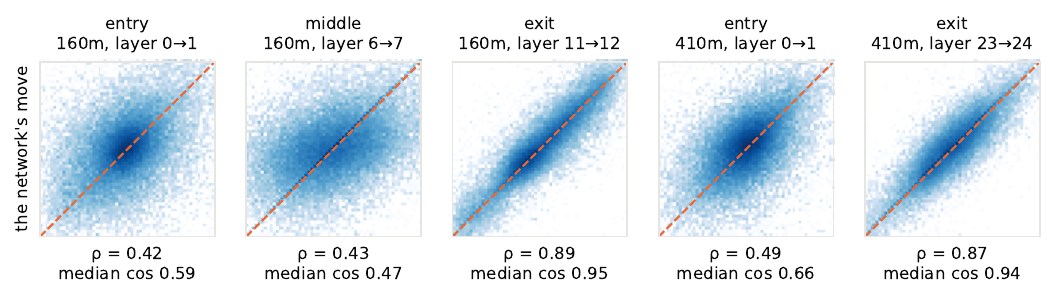}
\caption{For each token, the displacement prescribed by the exact optimal transport between the two layer clouds (horizontal) against the displacement the network actually applied (vertical), all 16 camera coordinates pooled, shown as a density. Agreement is the diagonal. $\rho$ is the rank agreement of the pooled components.}
\label{fig:agreement}\end{figure}

\begin{figure}[t]\centering
\includegraphics[width=\linewidth]{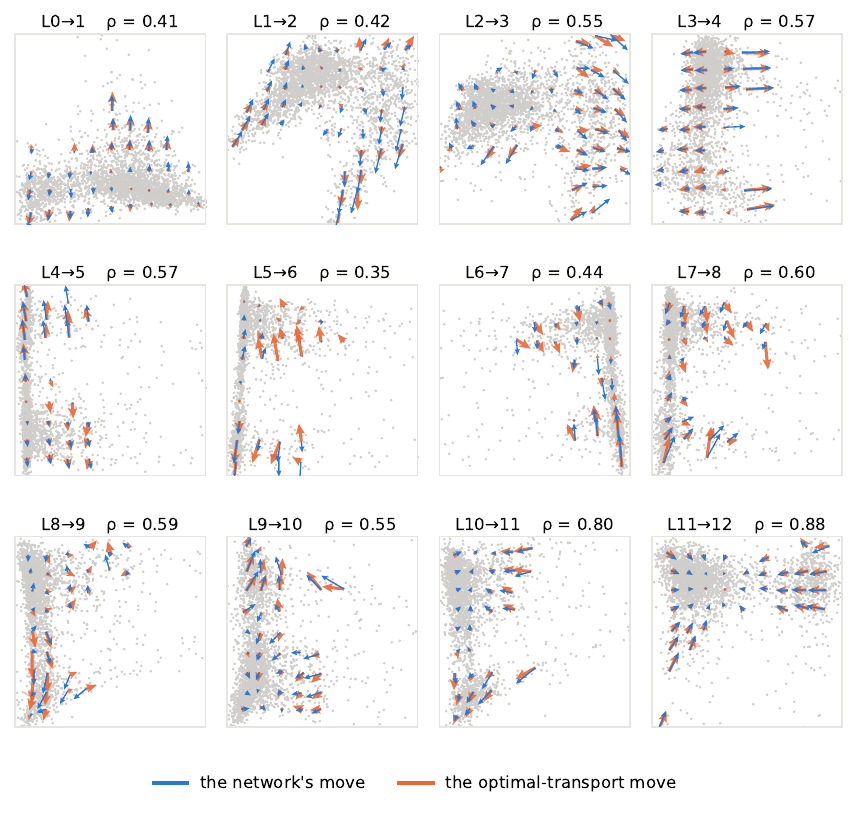}
\caption{For each transition of Pythia-160m, the network's move (blue) and the optimal-transport move (orange) for the token sample, drawn as bin means in the leading plane of the layer-$\ell$ cloud (16-d camera, centered clouds, exact assignment). One arrow scale per panel. $\rho$ is the rank agreement of the pooled 16-d components.}
\label{fig:fields}\end{figure}

We present the results in Table~\ref{tab:map}, including a null shuffle as control.
Agreement between the network's move and the optimal-transport move rises with depth (Figure~\ref{fig:profile}). The 410m curve dips near 0.4 of depth, at the transitions where its tokens move least. At the 160m exit the common shift is 99\% of the cost ($A/P$ 84 in the full space), so the map result concerns the remaining 1\%, which is resolvable (floor 0.11 of $P_\ell$). On Pythia-160m, $\rho$ is 0.41 at the entry, 0.38–0.62 in the middle, 0.80 at 10→11 and 0.89 at the exit, where the median cosine is 0.95 (Figure~\ref{fig:agreement}, and Figure~\ref{fig:fields} for all twelve transitions). Pile mean flows and raw states follow the same profile (exit $\rho$ 0.85 and 0.74). Pythia-410m ends at $\rho$ 0.88, median cosine 0.95. Seed ranges are ±0.01, and the shuffled-pairing null stays at $\rho\leq0.08$. The network's move is the optimal-transport move at the last layer, partly so in the middle, and barely at the first.

\paragraph{Is the agreement with the optimal-transport map more than a linear change?} To answer this, we fitted a linear optimal map, the Gaussian Brenier map $T_G$, from means and covariances (definition and full table in Appendix~\ref{app:affine}). It explains $39\%$ of the token-specific move at the 160m exit, and the remaining $61\%$ still agrees with the exact optimal transport ($\rho$ $0.83$, median cosine $0.94$, null $0.01$). At the 410m exit and at 160m $10\to11$ the linear map explains $90\%$ and $73\%$ and the remainder is below the floor. Elsewhere the linear map fits the network's moves at least as well as the full OT map and the remainder agrees at $\rho$ $0.27$--$0.36$.
In conclusion, the answer is yes for the 160m exit and no for every other transition we can resolve.

\paragraph{Is the exit agreement learned?} We repeat the measurement on Pythia-160m untrained (checkpoint \texttt{step0}), early in training (\texttt{step4000}) and fully trained, all from the same 5,000 documents. Exit rank agreement goes from $0.64$ to $0.67$ to $0.86$ (median cosine $0.74$, $0.75$, $0.94$), and at $L10\to11$ from $0.18$ to $0.42$ to $0.77$. At the entry the order reverses: $0.52$ untrained against $0.37$ trained. Therefore, the rise to $0.86$ is learned, and mostly after step $4000$. Training does not push the whole network toward optimal transport. It pushes the exit toward it (0.64 → 0.86) and the entry away from it (0.52 → 0.37). The linear control explains the $0.64$: at initialisation the exit move is linear (a linear map explains $97\%$ of it), and a linear map agrees with optimal transport by construction. Training replaces it by a move that is almost entirely non-linear, and that part follows optimal transport ($\rho$ $0.81$).


\begin{table}[t]\centering\small
\caption{Map agreement, 16-d camera, exact assignment on the centered clouds. $\rho$: rank agreement of the network's move with the optimal-transport move, per object; median cosine on WikiText; null $\rho$ from the shuffled pairing, the largest of the three objects. Seed ranges of $\rho$ are $\pm0.01$. Pythia-410m rows: WikiText only.}
\label{tab:map}
\begin{tabular}{l ccc c c}
\toprule
 & \multicolumn{3}{c}{$\rho$} & & \\
transition & WikiText & Pile & raw states & median cos & null $\rho$ \\
\midrule
0$\to$1 & 0.41 & 0.17 & 0.21 & 0.58 & 0.02 \\
1$\to$2 & 0.48 & 0.45 & 0.38 & 0.60 & 0.01 \\
2$\to$3 & 0.57 & 0.63 & 0.48 & 0.62 & 0.02 \\
3$\to$4 & 0.59 & 0.62 & 0.56 & 0.92 & 0.16 \\
4$\to$5 & 0.60 & 0.54 & 0.40 & 0.65 & 0.03 \\
5$\to$6 & 0.38 & 0.29 & 0.26 & 0.40 & 0.00 \\
6$\to$7 & 0.46 & 0.38 & 0.27 & 0.48 & 0.01 \\
7$\to$8 & 0.62 & 0.56 & 0.40 & 0.71 & 0.01 \\
8$\to$9 & 0.61 & 0.52 & 0.38 & 0.69 & 0.01 \\
9$\to$10 & 0.57 & 0.51 & 0.36 & 0.65 & 0.01 \\
10$\to$11 & 0.80 & 0.76 & 0.50 & 0.87 & 0.04 \\
11$\to$12 & 0.89 & 0.85 & 0.74 & 0.95 & 0.08 \\
\midrule
410m 0$\to$1 & 0.50 & -- & -- & 0.66 & 0.01 \\
410m 5$\to$6 & 0.59 & -- & -- & 0.97 & 0.10 \\
410m 6$\to$7 & 0.46 & -- & -- & 0.95 & 0.10 \\
410m 22$\to$23 & 0.70 & -- & -- & 0.86 & 0.09 \\
410m 23$\to$24 & 0.88 & -- & -- & 0.95 & 0.06 \\
\bottomrule
\end{tabular}

\end{table}

\section{Conclusion}\label{sec:conclusion}
 In this paper, we asked whether a trained transformer moves its token states the way optimal transport would. Two questions: does each layer pay the cheapest cost, and does each token go where the optimal map sends it? 
 To answer these questions, we compared the moves of Pythia-160m and Pythia-410m with exact optimal transport between consecutive layers, on two corpora, in a 16-dimensional projection. 


On cost, the token-specific move of the network is at the optimal reading at every transition we could resolve, except the last transition of Pythia-160m. On the map, the agreement between the network's move and the optimal-transport move increases with depth and reaches 0.89 at the last transition of Pythia-160m and 0.88 for Pythia-410m. This agreement is much weaker at initialisation and grows with training, while at the entry it decreases. For Pythia-160m it holds beyond any linear transformation of the cloud and for Pythia-410m it is accounted for by a linear map.

Our study suffers from several limitations, the main one being the resolution. With 4,000 tokens per sample, only the transitions where tokens move further than the sampling gap can be resolved: four of twelve in Pythia-160m and five of twenty-four in Pythia-410m. The middle layers could only be assessed in blocks of several layers. We can address this with larger sample and the required compute power. Another obvious limitation is that the study covers only two small models of one family on two small corpora. Finally, the mean flow averages each token over its contexts, which raises the agreement relative to raw states.

Despite these limitations, the picture that emerges is that the last layer of a trained transformer acts as an optimal-transport map on its cloud of states, the first layer does not, and the middle layers move each token too little to be judged one layer at a time. Several questions remain open. The middle layers may be linear, as the last layer of Pythia-410m is, but an instrument with finer resolution is needed to tell. The cost gap at the last layer of Pythia-160m, absent in Pythia-410m, may close with scale or may just be an artifact of Pythia's models. Finally, nothing in our results says whether the agreement with optimal transport is a property of language models or of trained transformers in general.
\bibliographystyle{plainnat}
\bibliography{references}

\appendix
\section{Linear map and remainder}
\label{app:affine}
This appendix gives the numbers behind the linear-map control of Section~\ref{sec:question_2}. The question is whether the agreement between the network's move and the optimal-transport move could come from a linear transformation of the whole cloud, a shift and a stretch, rather than from a token-by-token rearrangement.

The linear map is the Gaussian Brenier map $T_G$. For two clouds with means $m_\ell, m_{\ell+1}$ and covariances $\Sigma_\ell, \Sigma_{\ell+1}$, it is the optimal transport between the two Gaussians with those statistics,
\begin{equation}
T_G(x) \;=\; m_{\ell+1} + A\,(x - m_\ell), \qquad A \;=\; \Sigma_\ell^{-1/2}\big(\Sigma_\ell^{1/2}\Sigma_{\ell+1}\Sigma_\ell^{1/2}\big)^{1/2}\Sigma_\ell^{-1/2},
\end{equation}
with $A$ symmetric positive definite, so that $T_G$ is the gradient of a convex quadratic and hence a Brenier map. It maps the first cloud onto one with exactly the mean and covariance of the second, and it is fitted from those statistics alone. We measure two things. First, how much of the network's token-specific move the linear map explains, as the $R^2$ and the rank agreement $\rho$ between the linear move and the network's move. Second, what is left: we push the layer-$\ell$ cloud through $T_G$, take the move that remains from the pushed cloud to the target cloud, and compare it with the exact optimal transport between those two clouds, with the same floor and the same shuffled-pairing null as in the main text. The null reads $\rho\leq0.01$ at every transition.

Table~\ref{tab:affine} reports both. At the exit of Pythia-160m the linear map explains $39\%$ of the move and the remainder, $61\%$ of the cost, still agrees with optimal transport at $\rho$ $0.83$. At the exit of Pythia-410m and at $10\to11$ of Pythia-160m the linear map explains $90\%$ and $73\%$, and the remainder is smaller than its own floor, so its agreement cannot be measured. This last point matters for reading the table: a low remainder agreement next to a floor above one means "not resolvable", not "weak".

\begin{table}[h]\centering\scriptsize
\caption{Linear-map control, 16-d camera, $n=4{,}000$. Linear map: $R^2$ and $\rho$ of the move prescribed by $T_G$ against the network's token-specific move. Remainder: its share of the cost, its floor relative to its own cost, and its agreement with the exact optimal transport ($\rho$, median cosine). Last column: $\rho$ of the full optimal-transport map, for comparison.}
\label{tab:affine}
\begin{tabular}{l cc c c cc c}
\toprule
 & \multicolumn{2}{c}{linear map} & \multicolumn{4}{c}{remainder} & full map \\
transition & $R^2$ & $\rho$ & share of cost & floor & $\rho$ & median cos & $\rho$ \\
\midrule
160m 0$\to$1 & $-0.15$ & 0.35 & 1.15 & 0.07 & 0.27 & 0.30 & 0.40 \\
160m 1$\to$10 & 0.07--0.51 & 0.48--0.76 & 0.49--0.93 & 0.8--4.5 & 0.27--0.36 & 0.32--0.51 & 0.38--0.63 \\
160m 10$\to$11 & 0.73 & 0.81 & 0.27 & 0.47 & 0.44 & 0.52 & 0.80 \\
160m 11$\to$12 & 0.39 & 0.62 & 0.61 & 0.26 & \textbf{0.83} & \textbf{0.94} & 0.89 \\
raw states 11$\to$12 & 0.57 & 0.66 & 0.43 & 0.28 & 0.51 & 0.71 & 0.74 \\
410m 22$\to$23 & 0.96 & 0.82 & 0.04 & 2.3 & 0.35 & 0.36 & 0.70 \\
410m 23$\to$24 & 0.90 & 0.91 & 0.10 & 0.54 & 0.47 & 0.54 & 0.88 \\
\bottomrule
\end{tabular}
\end{table}

\end{document}